\documentclass[runningheads]{llncs}

\usepackage[T1]{fontenc}
\usepackage{graphicx}
\usepackage{placeins}
\usepackage{float}
\usepackage{amsmath,amssymb}
\usepackage{mathtools}
\usepackage{booktabs}
\usepackage[linesnumbered,ruled,vlined]{algorithm2e}
\usepackage{subcaption}
\usepackage{url}
\usepackage{hyperref}
\usepackage{color}

\graphicspath{{./}}

\begin{document}

%% ── Title & Anonymous Authors ─────────────────────────────────
\title{Variational Feature Compression for Model-Specific Representations}
\titlerunning{Variational Feature Compression for Model-Specific Representations}

\author{Zinan Guo\inst{1} \and Zihan Wang\inst{1} \and Chuan Yan\inst{1} \and Liuhuo Wan\inst{2} \and Zhiyong (Ethan) Ma\inst{1} \and Guangdong Bai\inst{3}}
\authorrunning{Z. Guo et al.}
\institute{The University of Queensland, Brisbane, Australia \\
\email {\{zinan.guo,zihan.wang\}@uq.edu.au, \{chuan.yan,zhiyong.ma\}@uqconnect.edu.au} \and
University of Wollongong, Wollongong, Australia \\
\email{liuhuow@uow.edu.au}\and
City University of Hong Kong, Hong Kong SAR, China\\
\email{g.bai@cityu.edu.hk}}

\maketitle

%% ── Sections ──────────────────────────────────────────────────
\begin{abstract}
As deep learning inference is increasingly deployed in shared and cloud-based settings, a growing concern is \emph{input repurposing}, in which data submitted for one task is reused by unauthorized models for another. Existing privacy defenses largely focus on restricting data access, but provide limited control over what downstream uses a released representation can still support. We propose a feature extraction framework that suppresses cross-model transfer while preserving accuracy for a designated classifier. The framework employs a variational latent bottleneck, trained with a task-driven cross-entropy objective and KL regularization but without any pixel-level reconstruction loss, to encode inputs into a compact latent space. A dynamic binary mask, computed from per-dimension KL divergence and gradient-based saliency with respect to the frozen target model, suppresses latent dimensions that are uninformative for the intended task. Because saliency computation requires gradient access, the encoder is trained in a \emph{white-box} setting, whereas inference requires only a forward pass through the frozen target model. On CIFAR-100, the processed representations retain strong utility for the designated classifier while reducing the accuracy of all unintended classifiers to below 2\%, yielding a suppression ratio exceeding 45$\times$ relative to unintended models. Preliminary experiments on CIFAR-10, Tiny ImageNet, and Pascal~VOC provide exploratory evidence that the approach extends across task settings, although further evaluation is needed to assess robustness against adaptive adversaries.

\end{abstract}

\section{Introduction}
\label{sec:intro}

The growth of Machine Learning as a Service (MLaaS) platforms has made it easier to deploy powerful models without requiring users to maintain their own infrastructure~\cite{papakostas2023mlaas,wang2024corelocker}. This convenience, however, comes with an important cost: users must often transmit sensitive inputs to remote servers for inference, making input data both the source of utility and a central point of exposure~\cite{zhang2021privacy,transparent24liu}.

A particularly important risk is the reuse of submitted inputs by unintended models. Even when a user provides data for a specific task, an adversary with access to the data path may apply the same input, or a processed version of it, to a different model for an unauthorized purpose. For example, a facial image submitted for gender classification could be repurposed for identity inference or emotion detection~\cite{wang2025nontransfer}. We refer to this threat as \emph{input repurposing}. Unlike membership inference or model inversion, input repurposing does not depend on access to the target model's parameters. Instead, it exploits the fact that a rich representation prepared for one task may still remain useful for others. Such repurposing can also conflict with principles of data minimization and informed consent~\cite{bouke2024leakage}.

Existing privacy-preserving techniques address related but different concerns. Homomorphic encryption~\cite{gilad2016cryptonets}, secure multi-party computation~\cite{bonawitz2017practical}, and differential privacy~\cite{sharma2020protecting} provide strong protection for data confidentiality or statistical disclosure, but they do not directly control whether a processed representation remains useful to unauthorized models. Cryptographic methods protect raw inputs during computation, yet any released output or transformed representation may still contain information that supports unintended downstream use. Differential privacy bounds what can be inferred about individual records, but it does not restrict the transferability of a released representation across models. These methods fundamentally address who can access the data; our focus, by contrast, is on what a released representation can still be used for.

\paragraph{Our Work.} We propose a feature-level transformation framework that suppresses cross-model transfer while preserving accuracy for a designated downstream classifier. The method uses a variational latent bottleneck, conceptually related to the Variational Information Bottleneck (VIB)~\cite{alemi2016deep}, to encode inputs into a compact latent space. Its guiding principle is to retain information useful for the intended prediction task while limiting information about the original input that may remain exploitable by other models. In practice, utility is encouraged through the cross-entropy loss of a frozen target model, while compression is imposed through KL divergence regularization toward a standard Gaussian prior. Unlike a standard VAE, no pixel-level reconstruction loss is used; the decoder is optimized to preserve target-relevant semantics rather than visual fidelity. A dynamic binary mask, computed from per-dimension KL divergence and gradient-based saliency, further suppresses latent dimensions that contribute little to the intended task.

We evaluate the framework on CIFAR-100 using four architectures (ResNet152, DenseNet121, ConvNeXt-V2, VGG16) as target and unintended models. With integrated dynamic masking, the designated classifier retains up to 72.23\% top-1 accuracy while all unintended models fall below 2\%, approaching the 1\% random baseline on 100 classes. Preliminary cross-dataset experiments on CIFAR-10, Tiny ImageNet, and Pascal~VOC confirm that selective utility generalizes beyond a single benchmark and task format.

Our main contributions are as follows:
\begin{itemize}
    \item A variational latent-bottleneck framework that learns task-driven reconstructions through the loss of a frozen target model, enforcing selective utility without a pixel-level reconstruction objective.
    \item A dynamic latent masking mechanism that combines KL divergence and gradient-based saliency to retain task-critical dimensions while suppressing transferable features.
    \item An empirical evaluation on CIFAR-100 across four architectures, including a three-stage ablation (no masking, KL-only, integrated) that isolates component contributions, showing target accuracy up to 72.23\% with unintended model accuracy below 2\%, together with exploratory cross-task experiments on CIFAR-10, Tiny ImageNet, and Pascal~VOC.
\end{itemize}

\paragraph{Scope and assumptions.}
Computing gradient-based saliency requires backpropagation through the frozen target classifier during training of the encoder and decoder. The method therefore assumes \emph{white-box} access to the target model at training time, while inference requires only a standard forward pass. This setting is realistic when the data holder operates the model directly or when the model is released with known weights, but it does not apply to a purely black-box MLaaS scenario in which only an inference API is available. We also emphasize that the proposed mechanism provides empirical transfer suppression rather than a formal privacy guarantee. Throughout the paper, we use the term \emph{selective utility} to denote the goal of maximizing accuracy for the designated model while reducing usefulness to others.

\section{Related Work}
\label{sec:related}

Our method lies at the intersection of privacy-preserving inference, feature-level representation filtering, and variational attribution methods. We briefly review these areas to clarify how the proposed framework differs from prior approaches and why cross-model transfer suppression is not directly addressed by existing methods.

\subsection{Input and Model-Level Privacy Defenses}

Input-level privacy techniques transform user data before transmission to remote models. Zhang et al.~\cite{zhang2019privacy} proposed distributed encoders that map inputs into compressed representations, reducing exposure while maintaining classification accuracy. Azizian and Baji\'{c}~\cite{azizian2024privacy} demonstrated that autoencoders can selectively preserve task-relevant features while removing private attributes. However, such approaches face the risk of partially reversible transformations~\cite{perez2024privacy} and often lack guarantees against feature reuse by unintended models.

Model-centric defenses modify the training process or model architecture. Differential privacy introduces statistical noise to provide quantifiable guarantees~\cite{prabhu2021privacy}, but often degrades accuracy on complex architectures~\cite{sharma2020protecting}. Cryptographic methods such as homomorphic encryption~\cite{gilad2016cryptonets} and secure multi-party computation~\cite{ruan2022private} enable inference over encrypted data but incur substantial computational costs, require architecture reconfiguration~\cite{kerschbaum2023privacy,wang2025rekey}, and face ongoing challenges in efficiently approximating non-linear activation functions~\cite{ma2025convexhull}. Neither input-level nor model-centric approaches directly address whether a processed representation remains exploitable by models other than the intended one.

\subsection{Feature Masking and Task-Specific Filtering}

Feature-level methods reshape intermediate representations to balance privacy and utility. Osia et al.~\cite{osia2020deep} proposed the Deep Private-Feature Extractor (DPFE), incorporating mutual information bounds to minimize sensitive content in extracted features. Ding et al.~\cite{ding2020privacy} introduced split-model architectures with client-side encoders, though intermediate representations may still encode sensitive attributes. Wang et al.~\cite{wang2023preserving} proposed Adaptive Feature Relevance Region Segmentation (AFRRS), which partitions features based on task correlation and applies targeted differential privacy noise.

Other feature masking approaches have explored selective transmission, task-agnostic anonymized representations, task-specific latent encoding, and region-level masking~\cite{alshammari2024privacy,li2020task,hajihassani2020latent,wei2024towards}. These methods show that feature filtering can improve the privacy-utility balance, but they generally focus on suppressing predefined sensitive attributes, injecting privacy noise, or learning reusable anonymized embeddings. They do not explicitly aim to suppress \emph{cross-model transfer}, that is, to make a processed representation useful to one designated model while reducing its utility for other models solving the same task. Complementary to model-side control methods such as AIM~\cite{wang2025aim}, our method learns an input-side transformation for model-specific selective utility.

\subsection{Variational Inference and Saliency Attribution}

The Information Bottleneck (IB) framework~\cite{tishby2000information} provides a theoretical foundation for extracting maximally relevant features while minimizing redundancy. The Variational Information Bottleneck (VIB)~\cite{alemi2016vib} extends this to deep learning, employing variational inference to balance compression and task fidelity. KL divergence serves as a proxy for the information cost of encoding, and recent work has demonstrated its effectiveness for latent space pruning by ranking feature importance~\cite{pan2023variable}. Notably, VIB differs from a standard VAE in that the training objective contains no pixel-level reconstruction likelihood; supervision comes entirely from the downstream task. These properties make VIB a natural foundation for selective-utility problems, where the goal is to retain task-specific information while discarding transferable content.

Saliency-based attribution methods such as GradCAM~\cite{selvaraju2017gradcam}, Integrated Gradients~\cite{sundararajan2017axiomatic}, and DeepLIFT~\cite{shrikumar2017deeplift} assign importance scores to features based on their contribution to predictions. Ahmidi~\cite{ahmidi2023latent} extended saliency analysis to latent representations, while region-based approaches like XRAI~\cite{kapishnikov2019xrai} demonstrated improved attribution quality. Most saliency methods operate on input-space features; applying them to latent representations requires gradient access to the downstream model, a constraint not shared by black-box approaches but one that enables finer-grained control over which latent dimensions are retained.

\section{Methodology}
\label{sec:method}

This section formalizes the threat model, introduces the proposed encoder--decoder architecture, and describes the variational bottleneck and saliency-guided masking mechanism used to enforce selective utility. We first specify the access assumptions and objective, then present the model components and training procedure.

\subsection{Threat Model}
\label{sec:threat}

We consider a setting in which a data holder wishes to perform inference using a specific, pre-trained classifier $f$ (the \emph{target model}). An adversary who can observe or intercept the processed input may attempt to feed it to a different model $g \neq f$ to extract information for an unauthorized task (\emph{input repurposing}).

\paragraph{Adversary capabilities.}
The adversary has access to the processed (reconstructed) image $X'$ and may evaluate it with any model of their choosing. The adversary does \emph{not} have access to the latent vector $Z$ or the mask $M$ directly. We do not assume an adaptive adversary who retrains a model specifically to invert the transformation; evaluating robustness against such adversaries is left to future work.

\paragraph{Defender capabilities.}
The data holder has \emph{white-box} access to the target model~$f$ during the training phase of the encoder and decoder (i.e., access to $f$'s architecture and weights for gradient computation). At inference time, only a forward pass through $f$ is required. This access profile is realistic when $f$ is an in-house model or a publicly released checkpoint with known weights.

\paragraph{Goal.}
The defender aims to learn a transformation $X \mapsto X'$ that maximizes classification accuracy under $f$ while minimizing the accuracy of any other model $g$ applied to $X'$. We call this property \emph{selective utility}. We do not claim formal privacy guarantees (e.g., differential privacy); the defense is empirical.

\subsection{Problem Definition}

Given an input image $X \in \mathbb{R}^{H \times W \times C}$ and its label $Y$, we seek a transformation
\begin{equation}
X' = \mathrm{Dec}(\mathrm{Enc}(X) \odot M)
\end{equation}
that produces a processed image $X'$ for downstream prediction by a designated model $f(\cdot)$. The goal is to preserve utility for the target model while limiting information that may remain useful to unintended models. Conceptually, this follows the Information Bottleneck principle, under which the encoder is encouraged to learn a latent representation $Z$ that retains information relevant to the label while compressing information about the input:
\begin{equation}
\max_{\mathrm{Enc}}  I(Z; Y) - \lambda \cdot I(Z; X)
\end{equation}
where $I(\cdot\,;\cdot)$ denotes mutual information and $\lambda$ controls the trade-off between utility and compression.

Because mutual information is not tractable to optimize directly, we adopt the VIB formulation~\cite{alemi2016vib}. In this setting, utility is promoted indirectly through a cross-entropy loss computed on the frozen target model, while compression is enforced through KL divergence between the encoder posterior $q(Z \mid X)$ and a standard Gaussian prior. The resulting tractable objective, $\mathcal{L}_{\text{task}} + \lambda_{\text{KL}} \cdot \mathcal{L}_{\text{KL}}$, is presented in Section~\ref{sec:vlb}. Throughout the remainder of this section, $Z \sim q(\cdot \mid X)$ denotes a latent sample drawn from the encoder posterior ($Z \in \mathbb{R}^d$, $d{=}512$; sampling details in Section~\ref{sec:vlb}), $M \in \{0,1\}^d$ denotes the binary mask, $Z_m = Z \odot M$ the masked latent vector, and $X' = \mathrm{Dec}(Z_m)$ the transformed image.

\subsection{System Architecture}

\begin{figure}[t]
\centering
\includegraphics[width=1\textwidth]{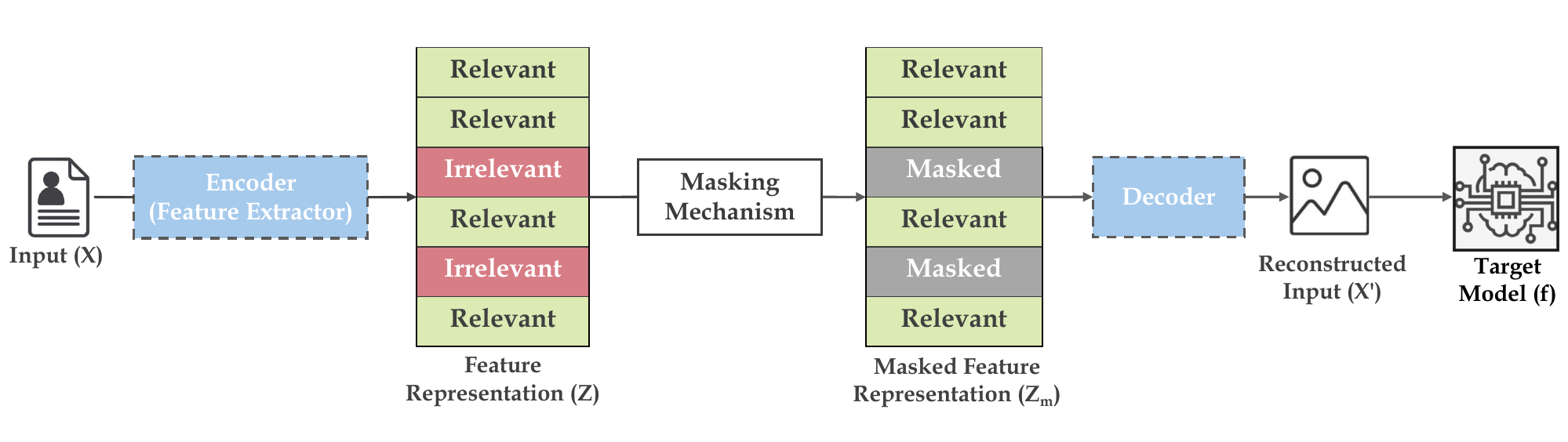}
\caption{Overview of the proposed selective-utility inference framework. An input image is encoded into a variational latent representation, filtered by a dynamic binary mask, decoded into a transformed image, and evaluated by a frozen target classifier. During training, gradients from the target-model loss update the encoder and decoder; at inference time, only a forward pass is required.}
\label{fig:framework_overview}
\end{figure}

Figure~\ref{fig:framework_overview} presents the overall framework. An input image $X$ is encoded into a latent representation $Z$. A masking mechanism filters $Z$ into $Z_m$, retaining only dimensions relevant to the designated task. The masked representation is decoded into $X'$, which serves as input to the frozen target model $f$ for inference. During training, gradients from $f$'s task loss flow back through the decoder and encoder (dashed arrows in the figure); at inference time only a forward pass through $f$ is needed.

\begin{figure}[t]
\centering
\includegraphics[width=1\textwidth]{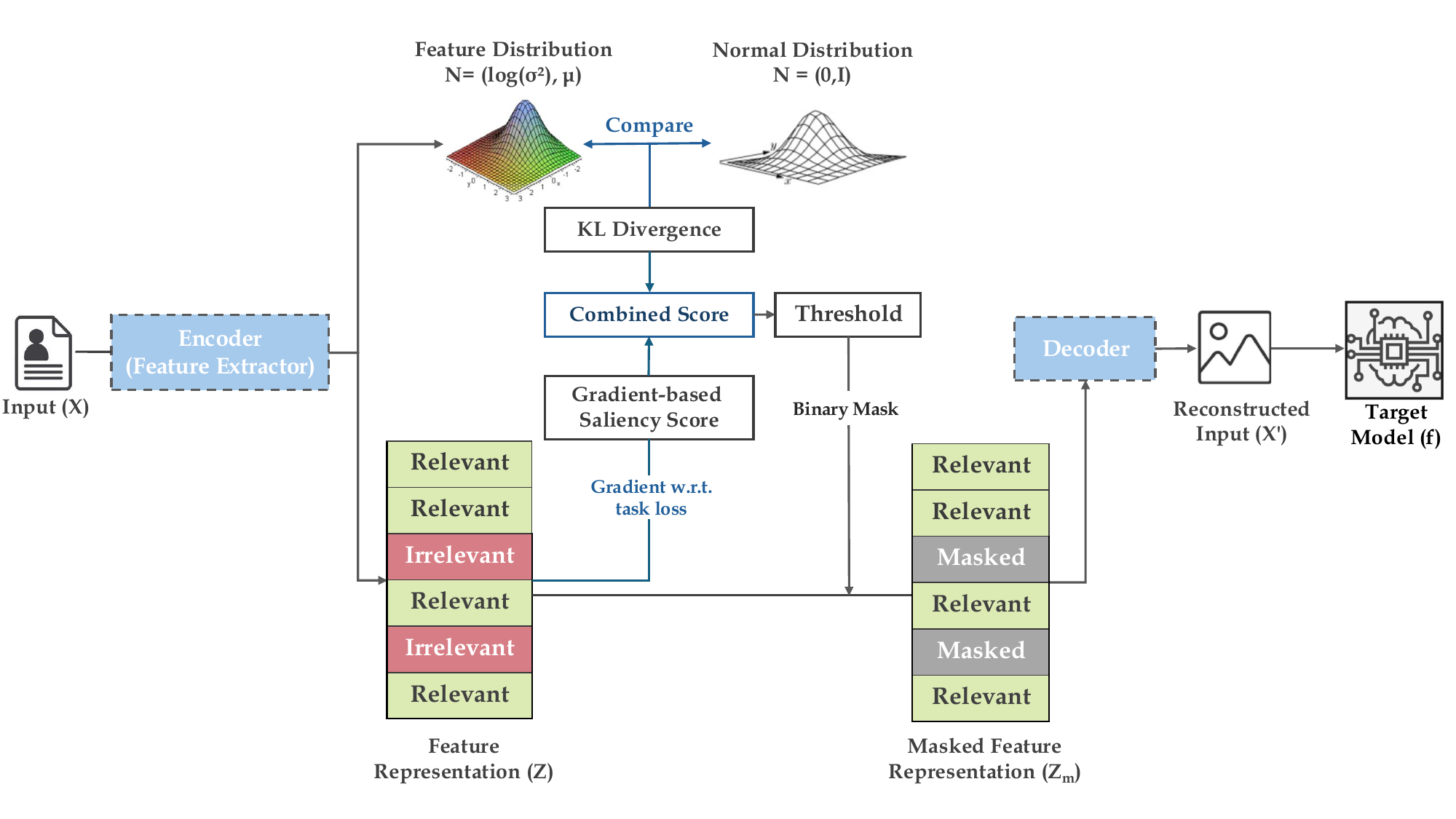}
\caption{Our feature filtering pipeline. The encoder produces latent features together with $\mu$ and $\log \sigma^2$. Each latent dimension is scored using KL divergence and gradient-based saliency, the scores are normalized and combined into a unified importance measure, and thresholding yields a binary mask that selects the dimensions retained for decoding.}
\label{fig:pipeline}
\end{figure}

Figure~\ref{fig:pipeline} details the feature filtering pipeline. The encoder outputs $Z$ along with distribution parameters $\mu$ and $\log\sigma^2$. Each dimension is scored by KL divergence and gradient-based saliency, normalized and combined into a unified importance score. A threshold produces a binary mask, and the masked vector $Z_m$ is decoded into $X'$ for inference.

\subsection{Variational Latent Bottleneck}
\label{sec:vlb}

\begin{figure}[t]
\centering
\includegraphics[width=0.4\textwidth]{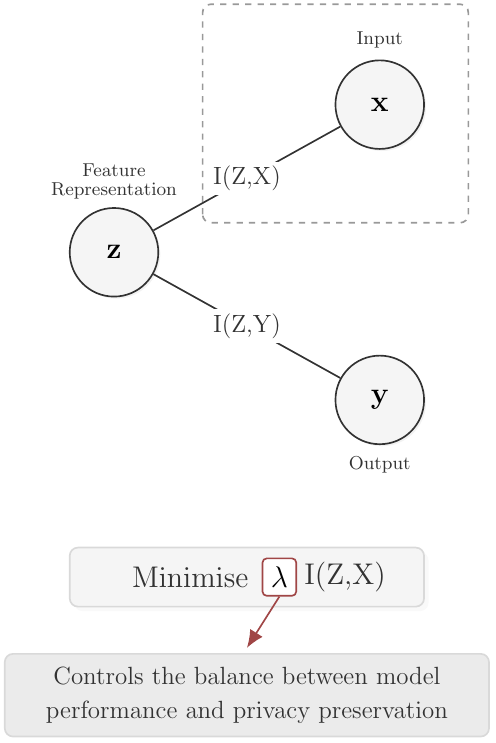}
\caption{Information Bottleneck view of the proposed variational latent bottleneck. The objective encourages the representation $Z$ to retain information useful for predicting $Y$ while suppressing information about the input $X$, with $\lambda$ controlling the compression-utility trade-off.}
\label{fig:vib}
\end{figure}

Our encoder uses a ResNet-18~\cite{he2016resnet} backbone with the final fully connected and softmax layers removed. The 512-dimensional output from global average pooling is processed through two parallel linear layers to produce $\mu$ and $\log\sigma^2$; a sample $Z$ is drawn via the reparameterization trick. Following the Variational Information Bottleneck (VIB) framework~\cite{alemi2016vib}, minimizing $I(Z;X)$ via KL divergence regularization reduces redundant information, while maximizing $I(Z;Y)$ through the task loss preserves predictive utility (Fig.~\ref{fig:vib}).

\paragraph{Distinction from a standard VAE.}
A standard VAE optimizes a reconstruction likelihood $\log p(X \mid Z)$. Our objective contains \emph{no} such term; the decoder is supervised entirely by the cross-entropy loss of the frozen target model. The system is therefore a VIB-style variational bottleneck with a task-driven decoder, not a generative model.

The training objective combines the target model's task loss with KL regularization:
\begin{equation}
\mathcal{L}_{\text{total}} = \mathcal{L}_{\text{task}} + \lambda_{\text{KL}} \cdot \mathcal{L}_{\text{KL}}
\end{equation}
where $\mathcal{L}_{\text{task}}$ is the cross-entropy loss computed by passing $X'$ through the frozen target model, and $\mathcal{L}_{\text{KL}}$ is:
\begin{equation}
\mathcal{L}_{\text{KL}} = \frac{1}{2} \sum_{i=1}^{d} \left( \mu_i^2 + \sigma_i^2 - \log \sigma_i^2 - 1 \right)
\label{eq:kl_loss}
\end{equation}
Dimensions with low KL divergence approximate the standard Gaussian prior $\mathcal{N}(0,I)$, indicating they encode little task-relevant information. The coefficient $\lambda_{\text{KL}}$ controls compression strength. Computing $\mathcal{L}_{\text{task}}$ requires backpropagating through the frozen target model; this is the white-box requirement described in Section~\ref{sec:threat}.

\subsection{Latent Masking Mechanism}

\begin{figure}[t]
\centering
\includegraphics[width=0.8\textwidth]{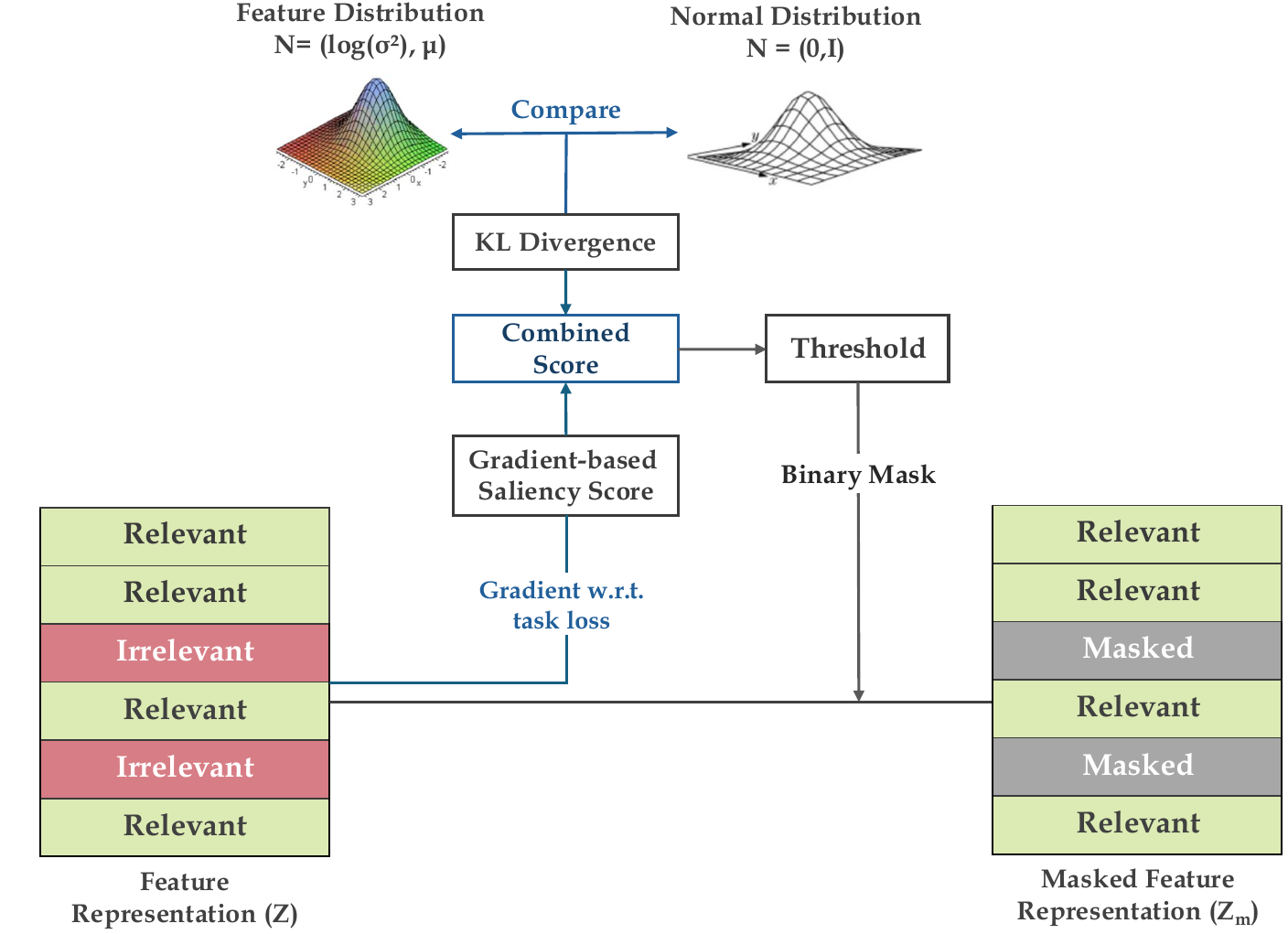}
\caption{The proposed latent masking mechanism. Per-dimension KL divergence and task-loss saliency are combined to estimate latent importance, and a thresholded binary mask suppresses dimensions that contribute little to the intended task.}
\label{fig:masking}
\end{figure}

KL regularization encourages compression, but it does not guarantee that all residual latent dimensions are irrelevant to unintended models. Some dimensions may remain weakly penalized by the bottleneck while still carrying information that supports transfer. To further refine the representation, we apply a post-hoc binary mask that combines two complementary notions of importance: statistical deviation from the prior and task relevance with respect to the frozen target model (Fig.~\ref{fig:masking}).

\paragraph{KL Divergence Score.} For each latent dimension $i$, we use the per-dimension KL contribution $\mathrm{KL}_i$ from Eq.~\eqref{eq:kl_loss}. Higher values indicate dimensions whose posterior deviates more strongly from the unit Gaussian prior and therefore carry more structured information. Lower values indicate dimensions that remain closer to the prior and contribute less to the encoded representation.

\paragraph{Gradient-Based Saliency Score.} KL divergence alone may not fully capture task relevance, since some dimensions can have modest KL values while still influencing the target prediction. To measure this effect, we compute the average absolute gradient of the task loss with respect to each latent dimension over a mini-batch:
\begin{equation}
S_i = \frac{1}{B} \sum_{b=1}^{B} \left| \frac{\partial \mathcal{L}_{\text{task}}}{\partial z_i^{(b)}} \right|.
\end{equation}
This score reflects the sensitivity of the frozen target model's loss to perturbations in latent dimension $i$. As with the task loss itself, computing this quantity requires gradient access to the target model during training.

\paragraph{Combined Score and Thresholding.} We min-max normalize both scores to $[0,1]$ and combine them as
\begin{equation}
I_i = \gamma \cdot \mathrm{KL}_i^{\mathrm{norm}} + (1-\gamma) \cdot S_i^{\mathrm{norm}},
\end{equation}
where $\gamma \in [0,1]$ balances statistical deviation from the prior against task relevance. The binary mask is then defined by
\begin{equation}
M_i =
\begin{cases}
1, & \text{if } I_i \geq T \cdot \max(I),\\
0, & \text{otherwise},
\end{cases}
\end{equation}
where $T$ is a threshold hyperparameter. Lower values of $T$ retain more latent dimensions and preserve utility for the target model, while higher values produce a sparser mask and stronger suppression. The resulting masked vector $Z_m = Z \odot M$ retains only high-importance dimensions.

\paragraph{Global Mask Aggregation.} The mask is computed globally over the training set and shared across samples. At each recomputation step, per-dimension KL scores are averaged over all training samples, while saliency scores are first averaged within each mini-batch and then averaged across batches. The resulting aggregate KL and saliency
vectors are normalized, combined, and thresholded to produce the global binary mask.

\subsection{Decoder}

After masking, the latent vector $Z_m \in \mathbb{R}^{512}$ is mapped back to image space by a decoder that produces the transformed input presented to the frozen target model. The decoder begins with a fully connected layer that projects $Z_m$ to a $3 \times 14 \times 14$ feature map. This representation is then progressively upsampled through four transposed-convolution blocks with stride~2 and kernel size~4, each followed by batch normalization and ReLU activation, yielding spatial resolutions of $14 \to 28 \to 56 \to 112 \to 224$. A final $\tanh$ activation constrains the output to the range $[-1,1]$.

The role of the decoder is not to reconstruct the original image as faithfully as possible, but to generate a transformed image that remains useful to the designated target classifier. For this reason, it is trained exclusively through the task loss induced by the frozen target model:
\begin{equation}
\mathcal{L}_{\text{task}} = \mathrm{CrossEntropy}\!\bigl(f(X'), Y\bigr).
\end{equation}
No pixel-level reconstruction objective, such as mean squared error or perceptual loss, is included. As a result, the decoder is encouraged to preserve decision-relevant structure rather than full visual fidelity. This design is deliberate: avoiding explicit reconstruction fidelity reduces the incentive to retain fine-grained details that could remain useful to unintended models.

\subsection{Training Strategy}

Training follows a two-phase schedule to avoid removing latent dimensions before their task relevance is reliably estimated. During warmup, the full latent vector $Z$ is decoded without masking, and the encoder and decoder are optimized under the VIB-style objective, $\mathcal{L}_{\text{task}} + \lambda_{\text{KL}} \cdot \mathcal{L}_{\text{KL}}$, while the target model remains frozen. This phase allows the encoder to form a stable task-relevant representation and gives the decoder time to learn how to produce transformed images that remain useful to the target classifier.

After warmup, training enters the masking phase. A global binary mask is computed from the combined KL--saliency score using statistics accumulated over the training set, and the mask is recomputed every \texttt{freq} epochs to reflect changes in feature importance as optimization progresses. Between recomputation steps, the mask is held fixed while the encoder and decoder continue to be updated using the same task and KL objectives. At inference time, the learned encoder, decoder, and final mask are applied directly: the input is encoded, filtered by the mask, decoded into a transformed image, and passed through the frozen target classifier using only a standard forward pass.

Algorithm~\ref{alg:training} summarizes the complete training procedure, highlighting the two-stage schedule, periodic mask recomputation, and mini-batch updates.

\vspace{10pt}

\begin{algorithm}[H]
\caption{Training with Saliency--KL Dynamic Masking}
\label{alg:training}
\KwIn{Training set $\mathcal{D}$, frozen target model $f$, encoder Enc, decoder Dec, hyperparameters $\gamma, T, \lambda_{\mathrm{KL}}, E_{\mathrm{warm}}, E_{\mathrm{mask}}, \mathrm{freq}$}
\KwOut{Trained parameters $\theta_{\mathrm{Enc}}, \theta_{\mathrm{Dec}}$}
$N \leftarrow E_{\mathrm{warm}} + E_{\mathrm{mask}}$
$M \leftarrow \mathbf{1}_d$ \tcp*{initialize mask to all-ones}
\For{$e = 1$ \KwTo $N$}{
  \tcp{Mask recomputation (masking phase only)}
  \If{$e > E_{\mathrm{warm}}$ \textbf{and} $(e - E_{\mathrm{warm}}) \bmod \mathrm{freq} = 0$}{
    $I_{\mathrm{KL}} \leftarrow \mathrm{GlobalKL}(\mathrm{Enc},\, \mathcal{D})$ \tcp*{mean per-dimension KL over samples}
    $I_{\mathrm{S}} \leftarrow \mathrm{GlobalSaliency}(\mathrm{Enc},\, \mathrm{Dec},\, f,\, \mathcal{D})$ \tcp*{mean per-dimension saliency over batches}
    $I \leftarrow \gamma \cdot \mathrm{MinMaxNorm}(I_{\mathrm{KL}}) + (1-\gamma) \cdot \mathrm{MinMaxNorm}(I_{\mathrm{S}})$
    $M_j \leftarrow \mathbf{1}[I_j \ge T \cdot \max(I)], \quad j = 1,\dots,d$
  }
  \tcp{Mini-batch updates}
  \ForEach{mini-batch $(X, Y) \subset \mathcal{D}$}{
    $\mu, \log\sigma^2 \leftarrow \mathrm{Enc}(X)$
    $\sigma \leftarrow \exp(\tfrac{1}{2}\log\sigma^2)$
    $Z \leftarrow \mu + \sigma \odot \epsilon, \quad \epsilon \sim \mathcal{N}(0, I)$ \tcp*{reparameterization}
    $X' \leftarrow \mathrm{Dec}(Z \odot M)$
    $\mathcal{L}_{\mathrm{task}} \leftarrow \mathrm{CrossEntropy}\!\bigl(f(\mathrm{Normalize}(X')), Y\bigr)$
    $\mathcal{L}_{\mathrm{KL}} \leftarrow
    \frac{1}{B}\sum_{b=1}^{B}\frac{1}{2}\sum_{j=1}^{d}
    \left((\mu_j^{(b)})^2 + (\sigma_j^{(b)})^2 - \log(\sigma_j^{(b)})^2 - 1\right)$
    $\mathcal{L} \leftarrow \mathcal{L}_{\mathrm{task}} + \lambda_{\mathrm{KL}} \cdot \mathcal{L}_{\mathrm{KL}}$
    Update $\theta_{\mathrm{Enc}}, \theta_{\mathrm{Dec}}$ via $\nabla_{\theta}\mathcal{L}$
  }
}
\end{algorithm}
\section{Experiments}
\label{sec:experiments}

We evaluate the proposed framework along three dimensions: target-model utility, suppression of transfer to unintended models, and generalization across datasets and task settings. Our primary study focuses on whether transformed inputs remain useful to the designated target classifier while becoming less useful to other pre-trained models applied to the same data. We then extend the evaluation to additional settings to examine whether the same mechanism remains effective beyond a single dataset or task formulation.

\begin{table}[t]
\centering
\caption{Baseline top-1 accuracy (\%) of the four pre-trained classifiers on original CIFAR-100 test images before any transformation is applied.}
\label{tab:cifar100-baseline}
\begin{tabular}{lc}
\toprule
\textbf{Model} & \textbf{Accuracy (\%)} \\
\midrule
VGG16 & 76.62 \\
ResNet152 & 85.46 \\
DenseNet121 & 83.61 \\
ConvNeXt-V2 & 88.34 \\
\bottomrule
\end{tabular}
\end{table}

\subsection{Experimental Setup}

All experiments are implemented in PyTorch and executed on a single NVIDIA RTX A6000 GPU (48\,GB). Optimization uses Adam with learning rate $1 \times 10^{-4}$ and weight decay $1 \times 10^{-5}$. Input images are resized to $224 \times 224$ and normalized using ImageNet statistics, and the batch size is fixed at 64 throughout. Unless otherwise noted, target models are frozen pre-trained checkpoints, and only the encoder and decoder are updated during training.

In the main integrated dynamic-masking experiments, we set $\gamma=0.5$ to give
equal weight to KL-based importance and target-model saliency, and update the
global mask every 5 epochs. Values of $\lambda_{\mathrm{KL}}$ and $T$ were selected on the validation split to maintain high target-model accuracy while keeping unintended-model accuracy near random.

We evaluate on four datasets covering a range of classification settings: \textbf{CIFAR-10}~\cite{krizhevsky2009learning} (10 classes, used for binary versus multiclass evaluation), \textbf{CIFAR-100}~\cite{krizhevsky2009learning} (100 fine-grained classes), \textbf{Tiny ImageNet}~\cite{tinyimagenet} (200 classes, with 64$\times$64 images resized to 224$\times$224), and \textbf{Pascal VOC 2012}~\cite{everingham2010pascal} (20 classes, multi-label). This combination allows us to test the proposed framework in standard multiclass classification, reduced-label settings, a cross-task transfer scenario, and a multi-label setting.

\subsection{CIFAR-100: Fine-Grained Classification}

Our primary evaluation of selective utility is conducted on CIFAR-100, where we test whether the transformed representation preserves utility for one designated classifier while suppressing transfer to other architectures trained for the same task. We use four pre-trained classifiers, ResNet152, DenseNet121, ConvNeXt-V2, and VGG16, with baseline accuracies on original test images reported in Table~\ref{tab:cifar100-baseline}. Each classifier serves as the target model in turn, supervising the encoder--decoder during training, while the remaining models are treated as unintended evaluators. This setup provides a controlled way to measure how strongly the transformed inputs remain tied to the designated model rather than supporting broad cross-model reuse.

\paragraph{Ablation Study.} We present results in three stages of increasing sophistication — no masking, KL-only masking, and integrated (KL + saliency) masking — to isolate the contribution of each component and to demonstrate that the observed suppression is not merely the result of information destruction but of targeted filtering.

\paragraph{Without Masking.} Training with task loss and KL divergence ($\lambda{=}0.01$) for 30 epochs without masking yields the results in Table~\ref{tab:cifar100-unmasked}. Target accuracy remains 67--71\%, while unintended accuracy stays near random (<3\%), showing that KL regularization alone provides baseline selective utility.

\begin{table}[t]
\centering
\caption{Top-1 accuracy (\%) on transformed CIFAR-100 test images produced without latent masking. Each column corresponds to a different target model used to supervise the encoder--decoder during training.}
\label{tab:cifar100-unmasked}
\setlength{\tabcolsep}{5pt}
\begin{tabular}{lcccc}
\toprule
& \textbf{VGG16} & \textbf{ResNet152} & \textbf{DenseNet121} & \textbf{ConvNeXt-V2} \\
\midrule
Target Accuracy & 67.77 & 67.44 & 68.57 & 70.60 \\
Mean Other Models & 1.14 & 1.23 & 1.21 & 1.08 \\
\bottomrule
\end{tabular}
\end{table}

\paragraph{KL-Only Dynamic Masking.} Using ResNet152 as target with dynamic mask updates every 5 epochs (15-epoch warmup + 30-epoch masking phase), KL-only masking achieves 68.60\% target accuracy while suppressing DenseNet121 to 1.00\% and ConvNeXt-V2 to 0.70\% at $\lambda{=}0.01$ (Table~\ref{tab:cifar100-kl}).

\begin{table}[t]
\centering
\caption{Top-1 accuracy (\%) on transformed CIFAR-100 test images under KL-only dynamic masking with ResNet152 as the target model. Lower off-target accuracy indicates stronger suppression of cross-model transfer.}
\label{tab:cifar100-kl}
\setlength{\tabcolsep}{5pt}
\begin{tabular}{cccc}
\toprule
$\lambda$ & \textbf{ResNet152 (Target)} & \textbf{DenseNet121} & \textbf{ConvNeXt-V2} \\
\midrule
0.01 & 68.60 & 1.00 & 0.70 \\
0.005 & 65.35 & 0.40 & 1.47 \\
\bottomrule
\end{tabular}
\end{table}

\begin{figure}[t]
\centering
\includegraphics[width=0.7\textwidth]{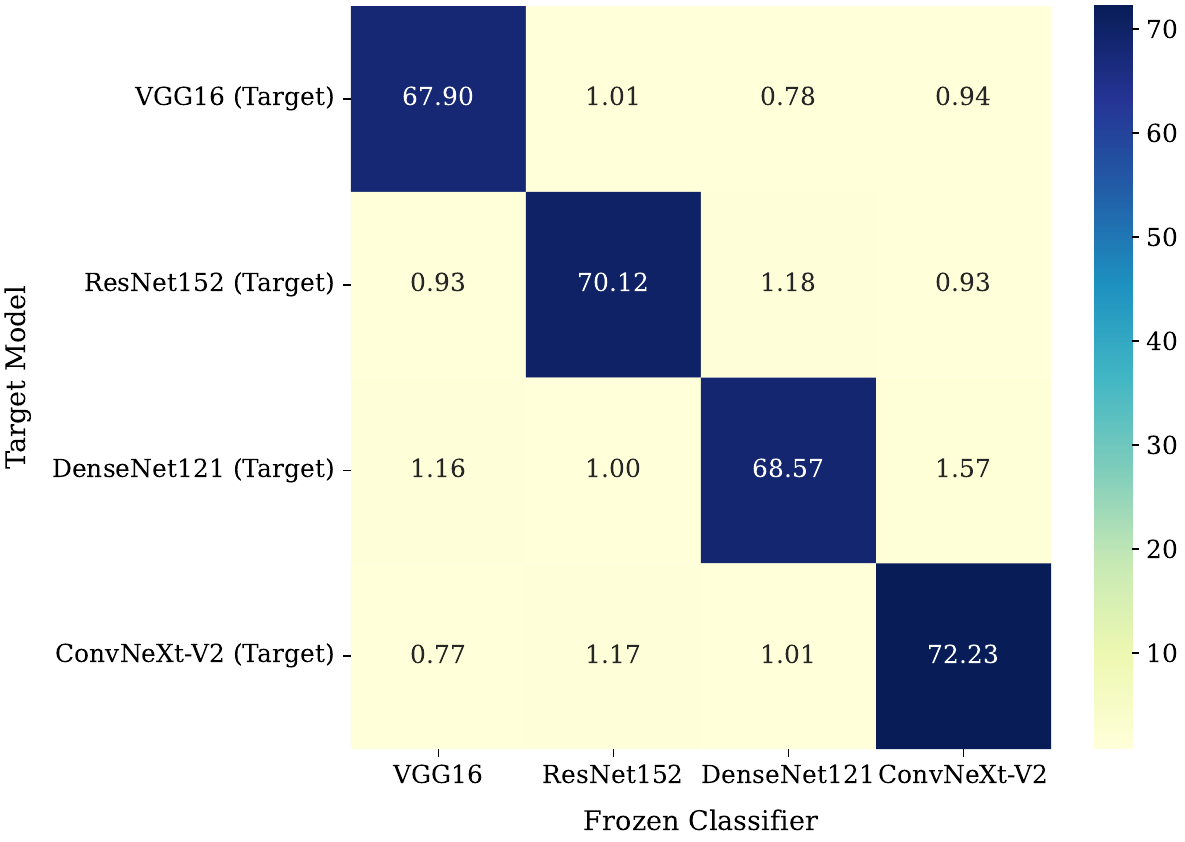}
\caption{Classification accuracy (\%) on CIFAR-100 under integrated dynamic masking. Diagonal entries show performance of the designated target model on transformed inputs, while off-diagonal entries show the accuracy of unintended classifiers applied to the same inputs. Strong diagonal contrast indicates selective utility and reduced cross-model transfer.}
\label{fig:heatmap}
\end{figure}

\paragraph{Integrated Dynamic Masking.} Combining KL divergence and saliency scores ($\gamma{=}0.5$) with a two-phase schedule (20-epoch warmup + 25-epoch masking, mask updates every 5 epochs) yields the best results.  As shown in Table~\ref{tab:cifar100-integrated} and Fig.~\ref{fig:heatmap}, integrated masking achieves up to 72.23\% target accuracy while keeping all unintended models below 1.6\%. Across the 12 off-diagonal entries, unintended accuracy averages 1.02\%, close to the 1\% random baseline on CIFAR-100. The best-performing configuration (ConvNeXt-V2 as target) yields a suppression ratio of approximately $46\times$ relative to the highest unintended accuracy in that row, indicating strong model-specific selectivity.

\paragraph{Ablation Summary.} Table~\ref{tab:ablation-summary} summarizes the progression across the three configurations, using ResNet152 as target. Each stage improves target accuracy while mean unintended accuracy remains near the 1\% random-guessing baseline, indicating that the masking components contribute additively rather than merely destroying information.

\begin{table}[t]
\centering
\caption{Ablation summary on CIFAR-100 with ResNet152 as target. Target accuracy improves across stages while mean unintended accuracy remains near chance (1\%).}
\label{tab:ablation-summary}
\setlength{\tabcolsep}{5pt}
\begin{tabular}{lcc}
\toprule
\textbf{Configuration} & \textbf{Target Acc.\ (\%)} & \textbf{Mean Unintended (\%)} \\
\midrule
No masking              & 67.44 & 1.23 \\
KL-only masking         & 68.60 & 0.85 \\
Integrated masking      & 70.12 & 1.01 \\
\bottomrule
\end{tabular}
\end{table}

\begin{table}[t]
\centering
\caption{Top-1 accuracy (\%) on transformed CIFAR-100 test images under integrated dynamic masking. Each row denotes the target model used during training, diagonal entries report target-model accuracy, and off-diagonal entries report the accuracy of unintended classifiers applied to the same transformed inputs.}
\label{tab:cifar100-integrated}
\setlength{\tabcolsep}{5pt}
\begin{tabular}{lcccc}
\toprule
\textbf{Target Model} & \textbf{ResNet152} & \textbf{DenseNet121} & \textbf{ConvNeXt-V2} & \textbf{VGG16} \\
\midrule
ResNet152   & \textbf{70.12} & 1.18 & 0.93 & 0.93 \\
DenseNet121 & 1.00 & \textbf{68.57} & 1.57 & 1.16 \\
ConvNeXt-V2 & 1.02 & 1.00 & \textbf{72.23} & 0.71 \\
VGG16       & 1.01 & 0.78 & 0.94 & \textbf{67.90} \\
\bottomrule
\end{tabular}
\end{table}

\subsection{Tiny ImageNet: Style Transfer Suppression}

This experiment tests cross-task privacy by using a ResNet-50 classifier as the target model and a pre-trained fast style transfer network~\cite{johnson2016perceptual} (applying Van Gogh's \textit{Starry Night} style) as the unintended model. Training uses integrated dynamic masking (25-epoch warmup + 35-epoch masking phase, mask updates every 5 epochs).

\begin{figure}[t]
\centering
\begin{subfigure}[b]{0.23\textwidth}
    \centering
    \includegraphics[width=\textwidth]{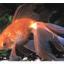}
    \caption{Original input}
\end{subfigure}
\hfill
\begin{subfigure}[b]{0.23\textwidth}
    \centering
    \includegraphics[width=\textwidth]{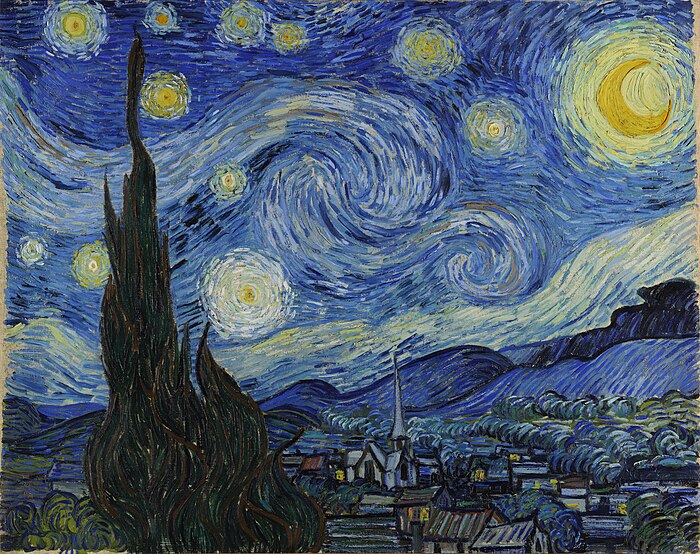}
    \caption{Style reference}
\end{subfigure}
\hfill
\begin{subfigure}[b]{0.23\textwidth}
    \centering
    \includegraphics[width=\textwidth]{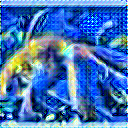}
    \caption{Stylized (ori.)}
\end{subfigure}
\hfill
\begin{subfigure}[b]{0.23\textwidth}
    \centering
    \includegraphics[width=\textwidth]{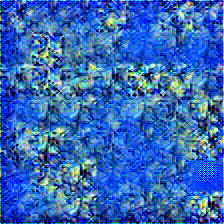}
    \caption{Stylized (recon.)}
\end{subfigure}

\caption{Qualitative cross-task transfer example. From left to right: original input, style reference, stylization of the original image, and stylization of the transformed image produced by the proposed framework. The degraded stylization result on the transformed image suggests reduced retention of features useful for this unintended downstream task.}
\label{fig:style}
\end{figure}

The target classifier achieves 60.81\% accuracy on reconstructed images (baseline: 74.25\% on originals), retaining reasonable classification performance. As shown in Fig.~\ref{fig:style}, stylized output from reconstructed images lacks semantic structure and clear content alignment, confirming that the learned representations discard generative information required for stylization while preserving classification-relevant semantics.

\subsection{Exploratory Extensions}

The following experiments provide preliminary evidence beyond CIFAR-100. We present them as exploratory rather than conclusive, since the evaluation covers only a subset of the full selective-utility protocol.

\paragraph{CIFAR-10: Binary vs.\ Multiclass.} Using static KL-only masking ($T{=}0.3$), the binary target task (living vs.\ non-living) reaches 96.75\% accuracy at $\lambda{=}0.05$, while the unintended 10-class classifier drops from 76.20\% to 17.64\%. However, we note that the random-guessing level for 10 classes is 10\%, so the remaining gap is less dramatic than in the 100-class setting. This suggests that separating coarse from fine-grained semantics is inherently easier than suppressing transfer among architectures trained on the same label space, and that the bottleneck may need to be tightened further in low-class-count settings.

\paragraph{Pascal VOC: Multi-Label Classification.} Using a ResNet-50 target model with static KL-only masking, the best configuration ($\lambda{=}0.01$, $T{=}0.3$) achieves 61.27\% Top-1 and 90.18\% Top-5 accuracy, demonstrating that the framework can preserve useful signal in a multi-label setting. Top-5 accuracy remains stable across configurations, indicating broad semantic retention even under stronger masking. We note that this experiment evaluates only target-model utility and does not yet measure whether unintended models are suppressed on the transformed VOC inputs; such evaluation is left to future work.

\section{Discussion}
\label{sec:discussion}

\subsection{Interpreting Selective Utility}

The most notable pattern in Fig.~\ref{fig:heatmap} is not simply that off-target accuracy is low, but that it remains low even when the unintended models solve the \emph{same} task on the \emph{same} dataset. This suggests that the method is not merely removing class information; rather, it is reshaping the input into a representation whose decision-supporting cues are aligned with one specific classifier and are no longer broadly reusable across architectures. In that sense, the reconstructed image should be viewed less as a faithful proxy for the original input and more as a \emph{target-conditioned surrogate} optimized for one downstream decision rule.

The progression from unmasked training to KL-only masking to integrated masking clarifies why this happens. KL regularization alone already discourages generic, high-capacity representations, which explains the strong suppression observed without an explicit mask. The saliency term then sharpens this bottleneck by preserving dimensions that matter to the target model's loss, even when those dimensions are not dominant under the KL criterion. The benefit of the integrated scheme is therefore not only stronger suppression, but a more selective form of retention: it keeps features because they are useful to the target, not because they are simply statistically active.

\subsection{The Main Trade-off}

The drop in target accuracy relative to unprocessed inputs is best understood as the price of making the representation less reusable. For example, ConvNeXt-V2 falls from 88.34\% on original CIFAR-100 images to 72.23\% after transformation. This gap is not just an optimization artefact; it is evidence that the bottleneck and mask are discarding visual structure that many models would otherwise exploit. Put differently, the same information that helps the target classifier reach its ceiling also appears to support transfer to other models, so suppressing reuse inevitably removes some of the target model's margin as well.

This interpretation has an important design implication. Increasing latent dimensionality, weakening the mask, or adding multi-scale decoding would likely recover some target accuracy, but such changes would also make the processed representation more natural and therefore more transferable. The key question is therefore not how to eliminate the accuracy gap entirely, but how to characterize the privacy--utility frontier explicitly. An important direction for future work is to map this frontier systematically, for example by reporting target accuracy against unintended-model accuracy as a function of bottleneck width, mask sparsity, or masking threshold, which would enable practitioners to select an operating point suited to their deployment constraints.

\subsection{Computational Cost}
The transformation pipeline has approximately 12.31M trainable parameters, mainly from the ResNet-18 encoder. Inference adds one encoder forward pass, one 512-dimensional mask operation, and one decoder forward pass, with no gradient computation; saliency estimation and mask recomputation are training-only. Since the transformation is target-specific, amortized reuse across related models is left to future work.

\subsection{Practical Scope and Open Limitations}

Several limitations follow directly from the way the method achieves selectivity. First, the defense is \emph{target-specific}: the encoder, decoder, and mask are trained against a particular frozen classifier, so changing the protected model may require retraining the transformation pipeline. This dependence supports model-specific selectivity, but limits claims of model-agnostic deployment.

Second, the current evaluation addresses \emph{off-the-shelf reuse} rather than worst-case privacy. The threat model excludes adaptive adversaries that retrain on transformed images, jointly optimize an attacker against the transformation, or attempt inversion. Such attacks may partially recover accuracy because the transformed images still retain target-discriminative structure. Until these evaluations are included, the results should be interpreted as empirical non-reusability under a closed set of unintended models rather than a general guarantee against downstream extraction.

Third, the evidence outside CIFAR-100 remains exploratory. CIFAR-10 suggests that coarse and fine semantics can be partially separated, Tiny ImageNet suggests degradation of generative content, and Pascal VOC shows that useful multi-label signal can be preserved. However, these experiments do not yet establish a general cross-task guarantee or evaluate unintended multi-label attackers.

Fourth, we do not compare directly against methods such as DPFE~\cite{osia2020deep} or adversarial feature extraction~\cite{ding2020privacy}, since these target different objectives (attribute suppression or statistical privacy) rather than model-specific selective utility. Adapting them to our evaluation protocol is non-trivial and is left to future work; our results instead establish an initial empirical reference point for the selective-utility setting.
\section{Conclusion}
\label{sec:conclusion}

We introduced a feature-level inference framework for reducing transferability to unintended models while preserving utility for a designated target classifier. The method combines a variational latent bottleneck with saliency-guided dynamic masking and trains a task-driven decoder through supervision from a frozen target model rather than pixel-level reconstruction. Across the evaluated settings, the framework maintains target-model utility while reducing unintended classifier accuracy to near-chance levels, suggesting that \emph{selective utility} is a practical objective when white-box access to the target model is available during training.

This study remains empirical and does not provide formal privacy guarantees. Future work will evaluate adaptive adversaries, broader tasks and architectures, and additional deployment settings.

%% ── References ────────────────────────────────────────────────
\bibliographystyle{splncs04}
\bibliography{References}

\end{document}